\documentclass[letterpaper, 10 pt, conference]{template/ieeeconf}  
\IEEEoverridecommandlockouts                              
\UseRawInputEncoding 
\usepackage{graphicx}
\graphicspath{{figures/}}
\usepackage{algorithm}
\usepackage{algpseudocode}
\usepackage{stfloats}

\usepackage{amsfonts} 
\usepackage{amssymb}  
\newcommand{\coloneqq}{\mathrel{\mathop:}=}

\usepackage{xcolor}
\usepackage{colortbl}
\usepackage{array}

\usepackage{amsmath, amssymb, bm}

\usepackage[separate-uncertainty = true]{siunitx}
\usepackage[hidelinks]{hyperref}
\usepackage[capitalise]{cleveref}
\usepackage{svg}
\usepackage{todonotes}
\usepackage{soul}

\usepackage{pgfplots}
\pgfplotsset{compat=1.18}
\usepackage{pgfplotstable}
\usepackage{float}
\usepackage{multirow}
\usepackage[utf8]{inputenc}
\usepackage[T1]{fontenc}

\usepackage{tikz}
\usetikzlibrary{arrows.meta, positioning, calc, fit, backgrounds, decorations.pathreplacing}

\renewcommand{\boldsymbol}[1]{\bm{#1}}

\usepackage{subcaption}
\usepackage{booktabs}

\Crefname{figure}{Fig.}{Fig.} 
\Crefname{section}{Sec.}{Sec.}
\Crefname{algorithm}{Alg.}{Alg.}
\Crefname{equation}{Eq.}{Eq.}
\creflabelformat{equation}{#2\textup{#1}#3}

\usepackage[
  backend=biber,        
  style=ieee,           
  doi=false,            
  isbn=false,           
  url=false,            
  eprint=true,          
  hyperref=true,        
  labeldateparts=false,
  minbibnames=3,
  maxbibnames=3,       
  sorting=none,         
  defernumbers=true,  
]{biblatex}
\usepackage{xpatch}
\AtEveryBibitem{%
\ifentrytype{book}{
    \clearfield{url}%
    \clearfield{urldate}%
    \clearfield{review}%
    \clearfield{volume}%
    \clearfield{number}%
    \clearfield{pages}%
    \clearfield{note}%
    \clearfield{address}%
    \clearfield{publisher}%
    \clearfield{language}%
    \clearfield{series}
}{}
\ifentrytype{collection}{
    \clearfield{url}%
    \clearfield{urldate}%
    \clearfield{review}%
    \clearfield{volume}%
    \clearfield{number}%
    \clearfield{pages}%
    \clearfield{language}%
    \clearfield{note}%
    \clearfield{publisher}%
    \clearfield{address}%
    \clearfield{series}
}{}
\ifentrytype{incollection}{
    \clearfield{url}%
    \clearfield{urldate}%
    \clearfield{review}%
    \clearfield{volume}%
    \clearfield{number}%
    \clearfield{pages}%
    \clearfield{note}%
    \clearfield{address}%
    \clearfield{publisher}%
    \clearfield{language}%
    \clearfield{series}
}{}
\ifentrytype{article}{
    \clearfield{url}%
    \clearfield{urldate}%
    \clearfield{note}%
    \clearfield{volume}%
    \clearfield{number}%
    \clearfield{pages}%
    \clearfield{note}%
    \clearfield{address}%
    \clearfield{publisher}%
    \clearfield{language}%
    \clearfield{extra}
}{}
\ifentrytype{misc}{
    \clearfield{url}%
    \clearfield{urldate}%
    \clearfield{volume}%
    \clearfield{number}%
    \clearfield{pages}%
    \clearfield{note}%
    \clearfield{address}%
    \clearfield{publisher}%
    \clearfield{language}%
}{}
\ifentrytype{inproceedings}{
    \clearfield{url}%
    \clearfield{urldate}%
    \clearfield{note}%
    \clearfield{extra}%
    \clearfield{volume}%
    \clearfield{number}%
    \clearfield{pages}%
    \clearfield{note}%
    \clearfield{address}%
    \clearfield{language}%
    \clearfield{publisher}%
    \clearfield{eventtitle}%
}{}
}

\IEEEaftertitletext{\vspace{-7pt}}

\title{\vspace{-0.7em}\LARGE \bf
A Unified Low-Dimensional Design Embedding for Joint Optimization of Shape, Material, and Actuation in Soft Robots
\vspace{-8pt}
}

\author{Vittorio Candiello$^{\mathbf{*} 1}$, Manuel Mekkattu$^{\mathbf{*} 1}$, Mike Y. Michelis$^{1,2}$, and Robert K. Katzschmann$^{1}$%
\thanks{$^{*}$ Equal contribution. Corresponding author: \href{mailto:rkk@ethz.ch}{\tt rkk@ethz.ch}.}
\thanks{$^{1}$ Soft Robotics Lab, D-MAVT, ETH Zurich, Switzerland.}%
\thanks{$^{2}$ ETH AI Center, ETH Zurich, Switzerland.}%
\vspace{-3pt}
}

\begin{document}

\maketitle
\thispagestyle{empty}
\pagestyle{empty}

\begin{abstract}
Soft robots achieve functionality through tight coupling among geometry, material composition, and actuation. As a result, effective design optimization requires these three aspects to be considered jointly rather than in isolation. This coupling is computationally challenging: nonlinear large-deformation mechanics increase simulation cost, while contact, collision handling, and non-smooth state transitions limit the applicability of standard gradient-based approaches. We introduce a smooth, low-dimensional design embedding for soft robots that unifies shape morphing, multi-material distribution, and actuation within a single structured parameter space. Shape variation is modeled through continuous deformation maps of a reference geometry, while material properties are encoded as spatial fields. Both are constructed from shared basis functions. This representation enables expressive co-design while drastically reducing the dimensionality of the search space. In our experiments, we show that design expressiveness increases with the number of basis functions, unlike comparable neural network encodings whose representational capacity does not scale predictably with parameter count. We further show that joint co-optimization of shape, material, and actuation using our unified embedding consistently outperforms sequential strategies. All experiments are performed independently of the underlying simulator, confirming compatibility with black-box simulation pipelines. Across multiple dynamic tasks, the proposed embedding surpasses neural network and voxel-based baseline parameterizations while using significantly fewer design parameters. Together, these findings demonstrate that structuring the design space itself enables efficient co-design of soft robots.
\end{abstract}

\begin{figure*}[b]
\vspace{-0.8em}
\centering
\includegraphics[width=\textwidth]{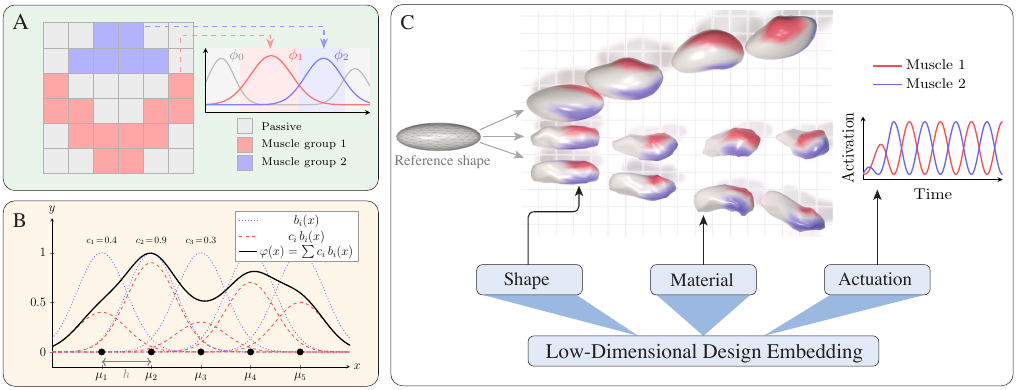}
\caption{\textbf{Overview of the Proposed Basis Function Design Embedding.} (A) Material distribution obtained by element-wise $\arg\max$ evaluation of the score field $\phi$, assigning each element to its dominant material phase. (B) Basis function parameterization: a finite set of spatial basis functions can approximate a continuous field. (C) Optimized swimmer trajectories (with lateral drift penalty): the basis function encoder with joint optimization (middle) produces a straight forward-moving trajectory, outperforming sequential optimization (bottom) and the neural field encoder (top), which exhibit a lateral drift.}
\label{fig:overview}
\end{figure*}


\section{Introduction}
The design of soft robots requires the simultaneous optimization of shape, material distribution, and actuation strategies to achieve complex dynamic behaviors. Unlike rigid robotic systems, soft robots exhibit highly nonlinear mechanics, large deformations, and often discontinuous material or activation models. These effects make simulations computationally expensive and significantly complicate design optimization~\cite{armanini_soft_2023}.

Classical approaches to design optimization, such as adjoint-based topology optimization, rely on analytic gradients of the objective function with respect to design variables~\cite{giannakoglou_adjoint_2008, allaire_structural_2004}. While highly effective in linear regimes, these methods require smooth governing equations and carefully derived adjoints. Consequently, the assumptions underlying adjoint-based optimization are difficult to satisfy in realistic soft-robot design pipelines. Similarly, differentiable-simulator approaches require end-to-end differentiability~\cite{hu_chainqueen_2019, du_diffpd_2021}, forcing all constitutive models, actuation laws, solvers, and integration schemes to admit derivatives with respect to state and design variables. In practice, this requirement significantly restricts the class of admissible physical models.

Realistic soft-robot simulations often include muscle-like actuation with saturation or state switching, contact and collision handling, and time-dependent events such as activation schedules or material transitions. These components typically rely on conditional logic or non-smooth updates that break differentiability or produce unstable gradients. Ensuring differentiability through such a pipeline requires substantial re-engineering and often forces algorithmic compromises that sacrifice physical fidelity or numerical robustness. 

Within this setting of a black box simulator, a key challenge becomes how to parameterize the design space. Geometry, material distribution, and actuation are in principle infinite-dimensional: shape is a spatial domain, material assignment a spatial field, and actuation a time-varying signal. Direct optimization is intractable without reducing the design to a finite-dimensional space. Naive element-wise encodings produce high-dimensional search spaces that scale poorly with mesh resolution, quickly becoming intractable for gradient-free optimization. In this work, we introduce a smooth, low-dimensional design embedding for soft robots. Our main contributions are:\\
\textbf{1) Unified low-dimensional design embedding using basis functions.} We propose a smooth, basis function-based design embedding that jointly parameterizes shape, multi-material distribution (\Cref{fig:overview}A), and actuation within a single parameter space, and derive its discretized formulation for direct implementation in numerical simulators.\\
\textbf{2) Controlled expressiveness with parameter efficiency.} We show that the representational capacity of our embedding scales predictably with the number of basis functions (\Cref{fig:overview}B), allowing explicit control over expressiveness while maintaining a compact parameterization.\\
\textbf{3) Empirical validation in black-box settings.} Across dynamic tasks involving large deformations and contact, we demonstrate that (i) joint co-optimization outperforms sequential strategies, and (ii) our embedding surpasses voxel-based and neural-network-based baselines (\Cref{fig:overview}C).

Overall, this work provides a general and flexible approach to soft-robot design optimization with an emphasis on the structure of the design space rather than the specifics of the optimization algorithm.

\section{Related Work}

\subsection{Topology and Shape Optimization}
Topology optimization (TO) methods traditionally represent geometry through implicit scalar fields, such as level-set or density formulations, and rely on adjoint-based derivatives to drive the optimization~\cite{giannakoglou_adjoint_2008}. Parametric level-set methods based on radial basis functions (RBFs)~\cite{wang_radial_2006, andrade_level-set-based_2022, jiang_topology_2020} reduce the dimensionality of the design space by expressing the level-set function as a linear combination of basis functions, leading to smooth boundaries and compact representations. While RBF-based level-set methods reduce dimensionality, they typically parameterize only geometry and remain embedded within adjoint-based optimization frameworks. In contrast, our approach extends basis function parameterization to jointly encode shape, multi-material distribution, and actuation within a single unified design space, explicitly targeting black-box simulation settings.

Extensions of TO to soft robots include nonlinear finite element methods (FEM), SIMP-based, and multi-material formulations~\cite{kumar_towards_2023, mehta_topology_2025, pinskier_diversity-based_2024, zhang_topology_2019}. Despite these advances, most TO-based methods remain tightly coupled to specific physical models, rely on mesh-dependent density fields, and depend on adjoint formulations that are difficult to extend to dynamic, contact-rich simulations. Surrogate-accelerated approaches such as ISMO~\cite{lye_iterative_2021} reduce cost but preserve these differentiability requirements, restricting applicability to complex soft-robot settings.

\subsection{Differentiable Simulation and Co-Design}
An alternative line of work focuses on making simulators differentiable, enabling gradient-based co-design of morphology and control~\cite{hu_chainqueen_2019, du_diffpd_2021}. DiffAqua~\cite{ma_diffaqua_2021} presents an end-to-end differentiable fluid–structure interaction framework for aquatic robots, allowing joint optimization of shape and actuation via automatic differentiation. By carefully constructing a differentiable simulator and using smooth shape parameterization (e.g., Wasserstein barycenters), such approaches avoid black-box optimization. While powerful, differentiable simulators impose strong constraints on the physical models, numerical solvers, and contact handling, which may require simplified constitutive laws. In contrast, our work does not modify the simulator or enforce differentiability, but instead focuses on structuring the design space itself to maintain effective optimization in black-box settings.

\subsection{Generative Encodings and Evolutionary Design}
In evolutionary robotics, generative encodings have been proposed to reduce the dimensionality of design spaces while preserving structural regularities~\cite{oh_deep_2019}. Compositional Pattern Producing Networks (CPPNs) map spatial coordinates to material or geometry labels through a single generative function, producing coherent morphologies from a low-dimensional parameter vector~\cite{stanley_compositional_2007}. From a conceptual standpoint, CPPNs and related implicit neural representations --- which we collectively refer to as \textit{neural field} encodings --- can be interpreted as functional design encodings, where geometry and material distributions are defined by a continuous field evaluated over space. Our approach shares this functional perspective but uses an explicit basis function parameterization. We will show that this explicit construction provides direct control over the expressiveness of the design space.

\section{Methods}

\begin{figure*}[t]
\vspace{-0.8em}
\centering
\input{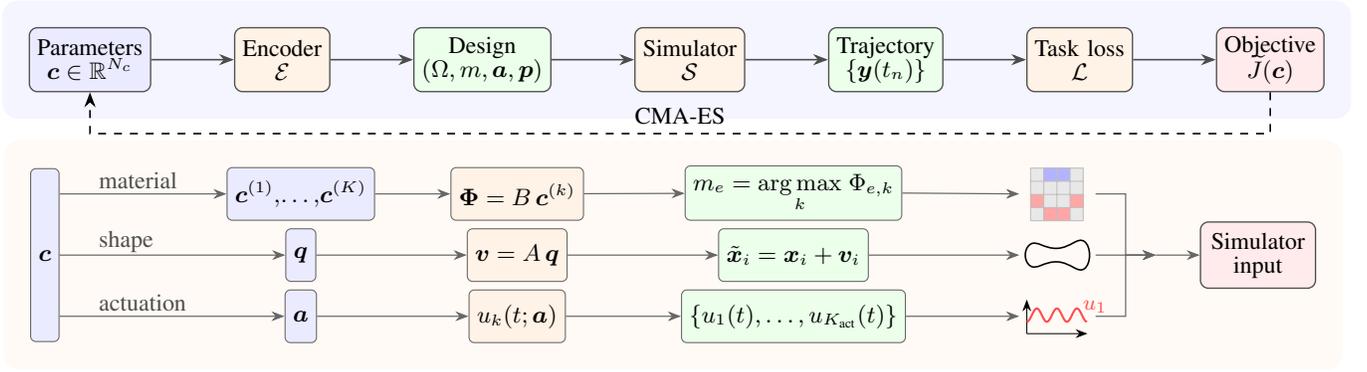}
\caption{\textbf{Co-design Optimization Pipeline.} The simulator evaluates a proposed design and computes its trajectory, CMA-ES updates parameter vector $\boldsymbol{c}$ to minimize the task loss (top). $\boldsymbol{c}$ is mapped by the encoder $\mathcal{E}$ to material fields, shape, and actuation signals via precomputed basis matrices $A$ and $B$ (bottom).}
\label{fig:architecture}
\vspace{-1.8em}
\end{figure*}

\subsection{Problem Formulation}
Let $\mathcal{D} \subset \mathbb{R}^d$ be a fixed design domain (bounding box) and let $[0,T]$ denote the simulation horizon. A soft-robot design is described by (i) a shape $\Omega \subset \mathcal{D}$, (ii) a spatial multi-material distribution, and (iii) time-dependent actuation signals associated with active components of the robot. Throughout this work, we refer to these active components as \emph{muscles}, although the formulation applies to arbitrary actuation mechanisms. We model the design through:
\begin{itemize}
    \setlength{\itemsep}{1pt}
    \setlength{\topsep}{0pt}
    \setlength{\parskip}{0pt}
    \setlength{\parsep}{0pt}
    \setlength{\partopsep}{0pt}
    \item a shape $\Omega$ (implicit set or deformation map),
    \item $K \in \mathbb{N}$ material fields $m_k:\mathcal{D}\to\mathbb{R}$ for each material or muscle group,
    \item actuation parameters $\boldsymbol{a} \in \mathbb{R}^{N_a}$, $N_a \in \mathbb{N}$ determining a function $u_a:[0,T]\to \mathbb{R}$ for each muscle group.
    \item external parameters $\boldsymbol{p} \in \mathbb{R}^{N_p}$, $N_p \in \mathbb{N}$ such as actuation direction or material stiffness.
\end{itemize}
Let $\boldsymbol{y}(t)$ denote the physical state of the system such as nodal positions and velocities. We consider a time-dependent soft body simulator, including contact and state-dependent constitutive laws:
\begin{align}
\mathcal{S}&: (\Omega, m, \boldsymbol{a}, \boldsymbol{p}) \mapsto \boldsymbol{y}_{0:N} = \{\boldsymbol{y}(t_n)\}_{n=0}^{N}, \\
t_n &= n\Delta t, \quad T=N\Delta t,
\end{align}
where $N\in \mathbb{N}$ is the number of timesteps. In practice, $\mathcal{S}$ is implemented as a variational FEM solver yielding a discrete trajectory of every node of the soft body. It may include non-smooth operations (contact, activation thresholds, behavior switching), which make reliable gradient computation impractical. We therefore treat $\mathcal{S}$ as an opaque black box.

Given a task-dependent loss $\mathcal{L}$ acting on the simulated trajectory, the design optimization problem is:
\begin{equation}
\min_{\Omega,\, m,\, \boldsymbol{a}, \boldsymbol{p}}~
\mathcal{L}\big(\{\boldsymbol{y}(t_n)\}_{n{=}0}^{N}\big) \text{ with }
\{\boldsymbol{y}(t_n)\}_{n{=}0}^{N} {=} \mathcal{S}(\Omega,m,\boldsymbol{a},\boldsymbol{p}).
\label{eq:problem_blackbox}
\end{equation}

Typical objectives include displacement, speed, jumping height, trajectory tracking, energy efficiency, or multi-term costs. Efficient optimization requires a compact design parameterization. We therefore seek a finite-dimensional parameterization of designs via a \emph{design encoder}
\begin{equation}
\mathcal{E} : \mathbb{R}^{N} \rightarrow \mathfrak{D},
\quad
\boldsymbol{c} \mapsto \mathcal{E}(\boldsymbol{c}) = (\Omega(\boldsymbol{c}), m(\boldsymbol{c}), \boldsymbol{a}(\boldsymbol{c})),
\label{eq:design_encoder}
\end{equation}
where $\boldsymbol{c} \in \mathbb{R}^{N}$ is a finite set of design parameters. The encoder $\mathcal{E}$ maps this vector to a concrete design, including shape, material distribution, and actuation. With this, \Cref{eq:problem_blackbox} becomes a tractable optimization problem in $\mathbb{R}^{N}$.

\subsection{Design Embedding}
\label{subsec:embedding}
We propose a smooth, low-dimensional basis function embedding of the design space. We first introduce the finite-dimensional basis function spaces underlying all components of the embedding. We then describe their use to encode multi-material distributions in continuous and discrete settings. Next, we present two mechanisms for shape parameterization: smooth morphing via deformation maps and topology changes through implicit occupancy fields. We then incorporate a parameterization of time-dependent actuation and additional physical parameters into the design vector. Continuous formulations are paired with their discrete implementation on a computational mesh. \Cref{fig:architecture} summarizes the optimization pipeline and the design embedding.

\textbf{Basis function spaces.} We define two finite-dimensional approximation spaces:
\begin{align}
\Phi &\coloneqq \mathrm{span}\{b_i\}_{i=1}^{N_\Phi} \subset C^0(\mathcal{D}), \\
\Psi &\coloneqq \mathrm{span}\{\boldsymbol{\psi}_j\}_{j=1}^{N_\Psi} \subset C^0(\mathcal{D};\mathbb{R}^d),
\end{align}
where $\{b_i\}$ are scalar basis functions and $\{\boldsymbol{\psi}_j\}$ vector-valued basis functions. In our implementation, we use Gaussian RBFs with fixed centers and widths. Their smoothness and spatial locality ensure that each parameter affects a bounded region of the domain, promoting coherent design patterns and avoiding high-frequency artifacts. Other choices of basis functions are possible and may offer different trade-offs between expressiveness and optimization difficulty.

\textbf{Multi-material encoding.} 
We encode $K$ material score fields $\phi_k(\cdot;\bm{c})\in \Phi$ for $K$ different materials:
\begin{equation}
\phi_k(\bm{x};\bm{c}) = \sum_{i=1}^{N_\Phi} c^{(k)}_i\, b_i(\bm{x}),
\qquad k=1,\dots,K,
\label{eq:phi_continuous}
\end{equation}
where $\boldsymbol{c}^{(k)}$ denotes the subvector of the full design vector $\boldsymbol{c}$ corresponding to the $k$-th material score field. In the default setting, material assignment is obtained through a hard selection rule. At each spatial location $\bm{x} \in \mathcal{D}$, the material index is defined as
\begin{equation}
m(\bm{x};\bm{c}) = \arg\max_{k \in \{1,\dots,K\}} \phi_k(\bm{x};\bm{c}).
\label{eq:hard_material_assignment}
\end{equation}
In practice, the design domain is discretized into $N_e$ cells with centers $\{\bm{x}_j\}_{j=1}^{N_e}$. All basis functions are pre-evaluated on this grid, yielding the basis matrix
\begin{equation}
B \in \mathbb{R}^{N_e \times N_\Phi},
\quad
B_{j,i} = b_i(\bm{x}_j).
\end{equation}
For each material $k$, let
\[
\bm{c}^{(k)} = (c^{(k)}_1,\dots,c^{(k)}_{N_\Phi})^\top \in \mathbb{R}^{N_\Phi}
\]
denote the vector of coefficients defining $\phi_k$. The discrete evaluation of the $k$-th material score field at the mesh element centers is obtained via the matrix--vector product
\begin{equation}
\boldsymbol{\phi}_k = B\, \bm{c}^{(k)} \in \mathbb{R}^{N_e},
\label{eq:phi_discrete}
\end{equation}
where $\boldsymbol{\phi}_k$ collects the values of $\phi_k(\bm{x},\bm{c})$ evaluated at $N_e$ element centers of the discretized domain. Collecting material scores of all $K$ materials at each element center yields
\begin{equation}
\Phi(\bm{c}) = \bigl[ \boldsymbol{\phi}_1 \;\; \boldsymbol{\phi}_2 \;\;
\dots \;\;
\boldsymbol{\phi}_K \bigr] \in \mathbb{R}^{N_e \times K},
\end{equation}
where each row corresponds to one grid cell and each column to one candidate material. Material assignment is then performed independently for each cell $j$ by
\begin{equation}
m_j = \arg\max_{k \in \{1,\dots,K\}} \Phi(\bm{c})_{j,k},
\end{equation}
which directly yields a voxel- or element-wise material labeling compatible with FEM pipelines. This multi-material encoding is illustrated in \Cref{fig:overview}A, while \Cref{fig:overview}B visualizes the approximation of a spatial field using basis functions.

Although the resulting material distribution is defined at the resolution of the computational grid, the number of design parameters is only $K N_\Phi$, which typically is orders of magnitude smaller than the number of cells $N_e$. Rather than assigning independent parameters to each voxel, this encoding groups cells implicitly through shared basis functions. As a result, the parameterization promotes spatial coherence in the material distribution. Complex material layouts can be represented by increasing the number and adapting the placement of the basis functions. 

For completeness, a smooth relaxation of the hard assignment in \Cref{eq:hard_material_assignment} can be obtained through a softmax operator with temperature $\tau > 0$:
\begin{equation}
w_k(\bm{x};\bm{c}) = \frac{\exp(\phi_k(\bm{x};\bm{c})/\tau)}
{\sum_{\ell=1}^{K} \exp(\phi_\ell(\bm{x};\bm{c})/\tau)},
\quad
\sum_{k=1}^{K} w_k(\bm{x};\bm{c}) = 1.
\label{eq:softmax_weights}
\end{equation}
In this case, material properties are blended continuously using the softmax weights, yielding a fully differentiable embedding suitable for gradient-based optimization. Both the hard and soft variants share the same basis function structure and differ only in the final assignment rule. The framework admits alternative basis functions, though their impact is not investigated here.

\textbf{Shape parameterization.} 
We support two complementary mechanisms for shape variation: \textit{(a) smooth morphing} of a reference shape via deformation maps, and \textit{(b) topology change via thresholding} of an implicit occupancy field. 
\paragraph*{(a) Morphing via deformation maps}
Let $\Omega_0 \subset \mathcal{D}$ be a reference shape (e.g., a cube or sphere) discretized by a mesh with fixed element connectivity. We parameterize a deformation field $\bm{v}(\cdot;\bm{c}) \in \Psi \subset C^0(\mathcal{D};\mathbb{R}^d)$ as
\begin{equation}
\bm{v}(\bm{x};\bm{c}) = \sum_{j=1}^{N_\Psi} q_j \, \boldsymbol{\psi}_j(\bm{x}),
\label{eq:deformation_field_continuous}
\end{equation}
where $\boldsymbol{q}$ denotes the subvector of $\boldsymbol{c}$ associated to the deformation field. The deformation map is defined as \begin{equation}
T_c(\bm{x}) = \bm{x} + \bm{v}(\bm{x};\bm{c}),
\label{eq:deformation_map}
\end{equation}
and the deformed shape is given by $\Omega(\bm{c}) {=} T_c(\Omega_0)$. In practice, the map $T_c$ is applied to mesh vertices, preserving element connectivity and yielding a continuous deformation of the original geometry.

Let $\{\bm{x}_i\}_{i=1}^{N_n}$ denote the mesh vertices of $\Omega_0$. All vector-valued basis functions are pre-evaluated at these nodes, producing the matrix
\begin{equation}
A \in \mathbb{R}^{(d N_n) \times N_\Psi},
\quad
A_{i \cdot \alpha,j} = \boldsymbol{\psi}_j^{(\alpha)}(\bm{x}_i),
\end{equation}
where $\alpha=1,\dots,d$ indexes spatial components. Stacking the nodal displacements yields
\begin{equation}
\boldsymbol{v} = A\, \bm{q} \in \mathbb{R}^{d N_n},
\label{eq:morphing_matrix_vector}
\end{equation}
which can be reshaped into displacement vectors for each node. Morphing the reference shape reduces to a single matrix--vector multiplication followed by a mesh update. To avoid inverted or degenerate elements, we restrict the deformation to small-strain regime via
\begin{equation}
\|\nabla \bm{v}(\cdot;\bm{c})\|_{\infty} \le \gamma < 1,
\end{equation}
which guarantees that we avoid overlapping nodes for $\gamma$ small enough. In practice, this can be implemented by bounding the coefficients $\bm{q}$ or scaling the displacement map $\boldsymbol{v}$. The morphing-based parameterization is differentiable with respect to the parameters $\bm{c}$ and preserves mesh topology.

\paragraph*{(b) Topology change via thresholding}
To allow for stronger topology modification, we additionally introduce an implicit occupancy field $\phi_{\Omega}(\cdot;\bm{c})\in\Phi$:
\begin{equation}
\phi_{\Omega}(\bm{x};\bm{c}) = \sum_{i=1}^{N_\Phi} c_{\Omega,i}\, b_i(\bm{x}),
\label{eq:occupancy_field}
\end{equation}
where $\boldsymbol{c}_\Omega$ is the subvector of $\boldsymbol{c}$ corresponding to the occupancy field, and $\{b_i\}$ are scalar basis functions, consistent with the material encoding. The shape is then defined implicitly as
\begin{equation}
\Omega(\bm{c}) = \{ \bm{x} \in \mathcal{D} \mid \phi_{\Omega}(\bm{x};\bm{c}) \ge \theta \},
\label{eq:threshold_shape}
\end{equation}
for a fixed threshold $\theta \in \mathbb{R}$.
On an element-based discretization with centers $\{\bm{x}_j\}_{j=1}^{N_e}$, the occupancy field is given by
\begin{equation}
\boldsymbol{\phi}_{\Omega} = B\, \bm{c}_{\Omega} \in \mathbb{R}^{N_e},
\end{equation}
using the same basis matrix $B$ introduced for material encoding. Elements are retained or removed according to
\begin{equation}
\text{element } j \in \Omega(\bm{c})
\quad \Longleftrightarrow \quad
(\boldsymbol{\phi}_{\Omega})_j \ge \theta.
\end{equation}
To further reduce the number of design parameters, the occupancy field $\phi_{\Omega}$ can be constructed directly from the material score fields introduced in the previous section. In particular, instead of introducing an independent set of coefficients
$\bm{c}_{\Omega}$, we define
\begin{equation}
\phi_{\Omega}(\bm{x};\bm{c}) \coloneqq \sum_{k=1}^{K} \phi_k(\bm{x};\bm{c}),
\label{eq:shape_from_materials}
\end{equation}
where $\{\phi_k\}_{k=1}^K$ are the material score fields. This construction follows from viewing the material score field as an indicator of material presence. At each location, if the sum of all score fields exceeds threshold $\theta$, the space is considered occupied. This construction implicitly couples shape and material distribution. While this reduces expressiveness compared to an independent shape field, it significantly lowers the design space dimension and simplifies the optimization.

This mechanism enables topology changes through simple element-wise evaluation and thresholding. However, the resulting parameter-to-geometry map is non-smooth due to discrete element addition and removal. It is worth noting that this non-smoothness is mesh-dependent: with finer discretization, the continuous field $\phi_\Omega(\cdot;\bm{c})$ is sampled more densely, and geometric variations become smoother at the macroscopic scale. In contrast, morphing provides a smooth, topology-preserving deformation that is fully differentiable. Moreover, thresholded designs often result in overly idealized or simplified geometries, which may be less representative of realistic manufacturable structures compared to smoothly morphed configurations.

\textbf{Actuation encoding.} 
While the primary focus of this work is on the design embedding of morphology and material distribution, co-optimization of soft robots additionally requires time-dependent actuation. We do not address feedback control or policy learning in this work. Actuation is treated as a parameterized open-loop signal. This choice allows us to study co-design problems where morphology, material distribution, and actuation parameters are optimized simultaneously, without introducing additional control complexity. Each active material group is associated with an independent parameterized actuation function, allowing different regions of the robot to be driven by distinct temporal patterns. Possible options include discrete-time actuation schedules, linear combinations of temporal basis functions (e.g. Gaussian pulses), or harmonic representations using sine and cosine functions. All these choices yield a finite-dimensional actuation parameter vector that can be handled uniformly by the optimization procedure.

\textbf{Additional parameters.} 
The proposed framework allows the inclusion of additional design parameters beyond shape, material assignment, and actuation. Such parameters may include stiffness coefficients, actuation directions, density, or other problem-dependent physical or control variables. These parameters are appended to the design vector and are optimized jointly with the geometric and material coefficients, without requiring changes to the simulation pipeline or the design embedding itself.

\textbf{Complete finite-dimensional design embedding.}
Collecting all parameters into $\bm{c}\in\mathbb{R}^{N}$, the embedding map
\begin{equation}
\mathcal{E}(\bm{c}) = (\Omega(\bm{c}), \mathcal{M}(\cdot;\bm{c}), u_a(\cdot;\bm{c}))
\end{equation}
induces a finite-dimensional manifold of admissible designs
\begin{equation}
\mathcal{M} \coloneqq \{\mathcal{E}(\bm{c})\mid \bm{c}\in\mathbb{R}^{N}\}
\end{equation}
embedded in the infinite-dimensional space of shapes and fields. Under mild assumptions on the placement and scaling of the Gaussian RBFs, the spans $\Phi$ and $\Psi$ are dense in $C(\mathcal{D})$ and $C(\mathcal{D};\mathbb{R}^d)$~\cite{buhmann_radial_2000}. Hence, increasing $N_\Phi$ and $N_\Psi$ increases expressiveness, while moderate choices keep optimization tractable. Our experimental results empirically support this scaling behavior. 

\subsection{Optimization}
Using the proposed design embedding, the original co-design problem reduces to a finite-dimensional optimization problem over the parameter vector $\bm{c} \in \mathbb{R}^N$:
\begin{equation}
\min_{\bm{c} \in \mathbb{R}^N} \; \tilde{J}(\bm{c}),
\quad
\tilde{J}(\bm{c}) \coloneqq J(\mathcal{E}(\bm{c})) = \mathcal{L}\big(\mathcal{S}(\mathcal{E}(\bm{c}))\big),
\label{eq:reduced_problem}
\end{equation}
where $\mathcal{E}(\bm{c})$ denotes the decoding map from parameters to a concrete design, $\mathcal{S}$ the forward simulation, and $\mathcal{L}$ the task-specific loss. We solve~\Cref{eq:reduced_problem} using the Covariance Matrix Adaptation Evolution Strategy (CMA-ES), a population-based, gradient-free optimization method. CMA-ES is well suited for this setting because the objective function is non-convex, potentially noisy, and evaluated through expensive time-dependent simulations. We favor CMA-ES over Bayesian optimization due to its superior scalability in higher-dimensional design spaces~\cite{santoni_comparison_2024}. CMA-ES operates directly on objective evaluations without gradient or surrogate models, and all candidates within a generation can be evaluated independently in parallel --- a critical advantage when each evaluation requires a full forward simulation. At each iteration, CMA-ES samples a population of candidate solutions from a multivariate normal distribution:
\begin{equation}
\bm{c}^{(i)} \sim \mathcal{N}\big(m,\, \sigma^2 C\big), \quad i = 1,\dots,\lambda,
\end{equation}
where the mean $m$, covariance matrix $C$, and global step size $\sigma$ are updated based on the ranking of candidates according to their objective values.
While we instantiate the framework using CMA-ES, the proposed design embedding is not tied to a specific optimizer. Other gradient-free methods can be used without modification, and where differentiable simulators are available, our smooth parameterization can also be combined with gradient-based optimization techniques.

\section{Results}

The experimental evaluation investigates three key aspects: (i) the expressiveness of the proposed design embedding, (ii) the importance of co-designing shape, material, and actuation, and (iii) the practical advantages of the embedding compared to baseline encodings such as implicit neural fields or voxel-based parameterization. All experiments use the same black-box simulation pipeline~\cite{mekkattu_sors_2025}, optimizer, and evaluation budget, unless otherwise specified.

\subsection{Expressiveness of the Design Embedding}
We first study the expressive power of the proposed encoding independently of task-level dynamics. In particular, we evaluate how the representational capacity scales with the number of basis functions, both for material distribution and shape variation.

\begin{figure}[b]
  \centering
  \vspace{-1em}
  \setlength{\tabcolsep}{1pt}
  \renewcommand{\arraystretch}{0.8}
  \begin{tabular}{cccc}
    \scriptsize $N_\Phi{=}16$ & 
    \scriptsize $N_\Phi{=}36$ & 
    \scriptsize $N_\Phi{=}64$ &
    \scriptsize Target \\
    \includegraphics[width=0.23\columnwidth]{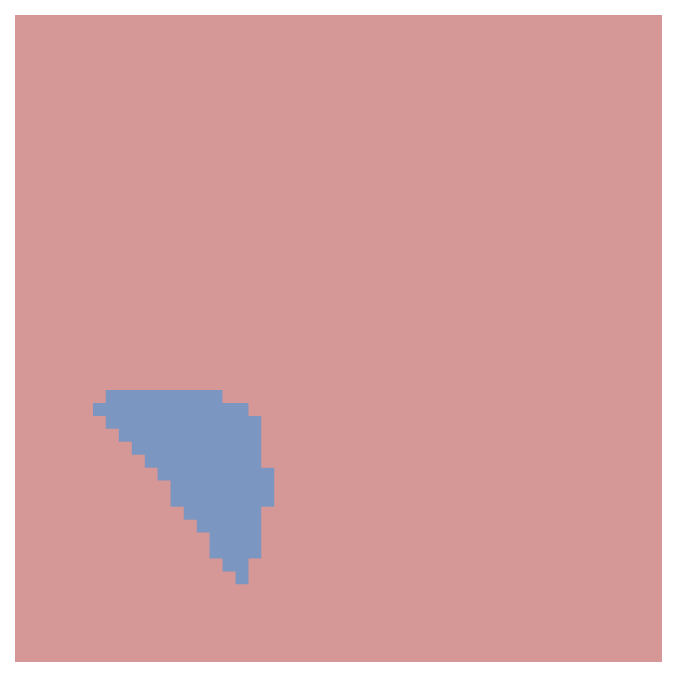} &
    \includegraphics[width=0.23\columnwidth]{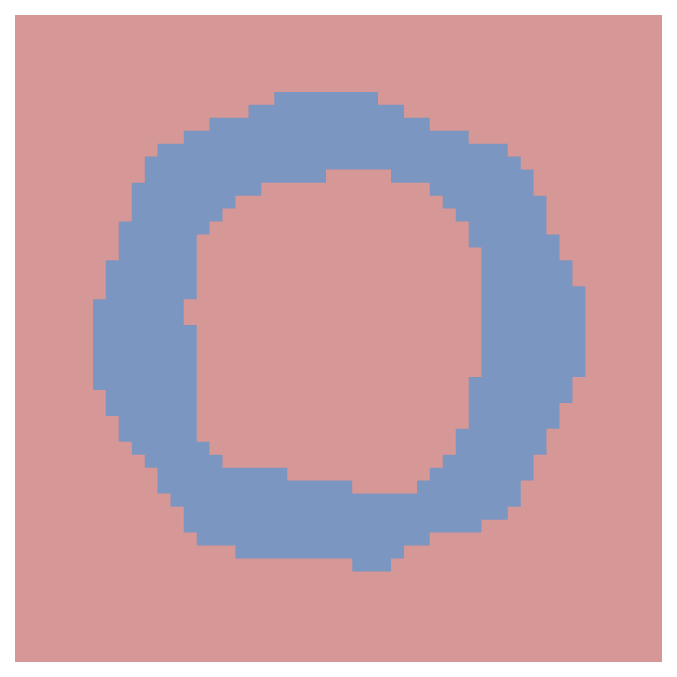} &
    \includegraphics[width=0.23\columnwidth]{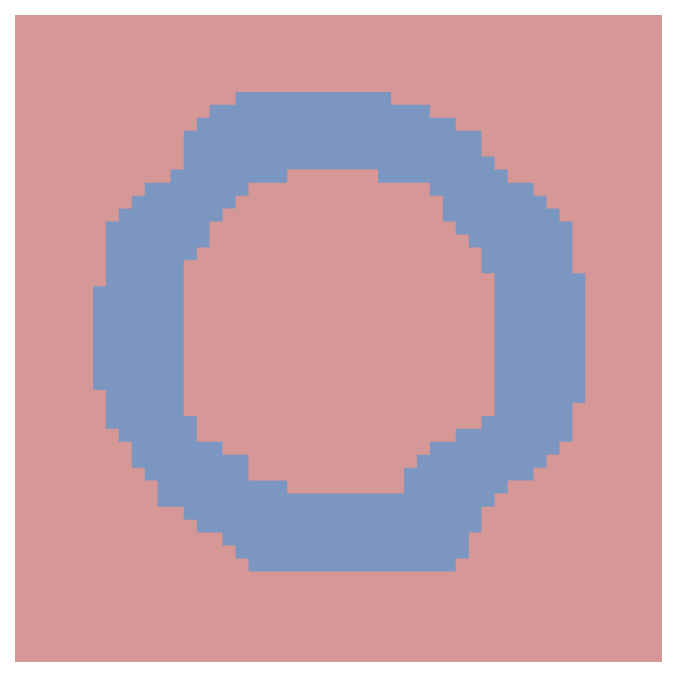} &
    \fbox{\includegraphics[width=0.23\columnwidth]{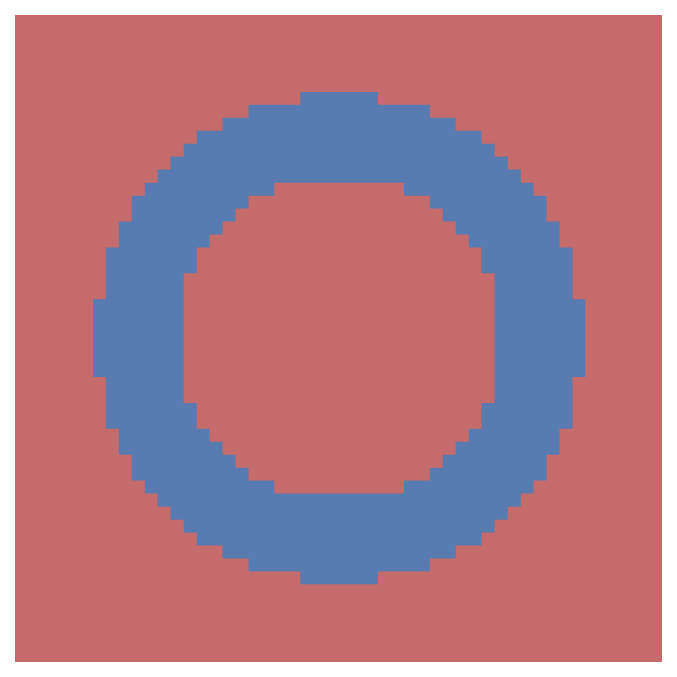}} \\
    \includegraphics[width=0.23\columnwidth]{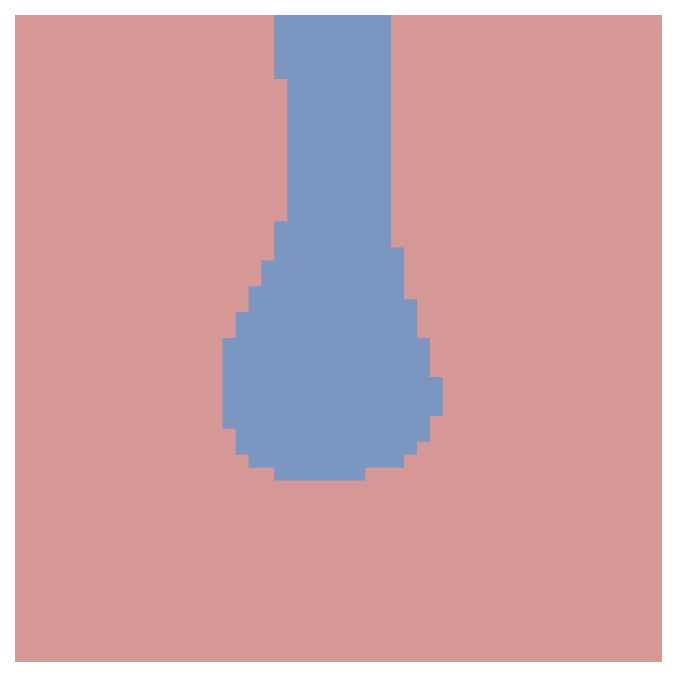} &
    \includegraphics[width=0.23\columnwidth]{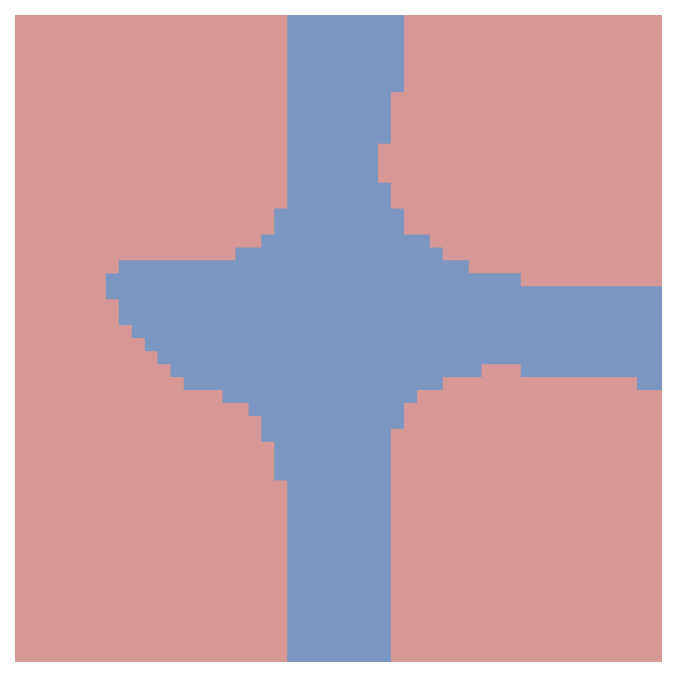} &
    \includegraphics[width=0.23\columnwidth]{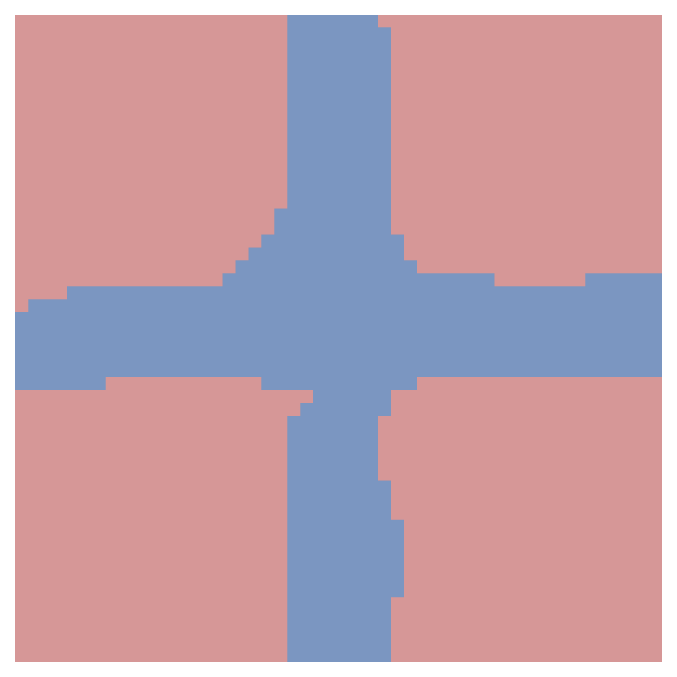} &
    \fbox{\includegraphics[width=0.23\columnwidth]{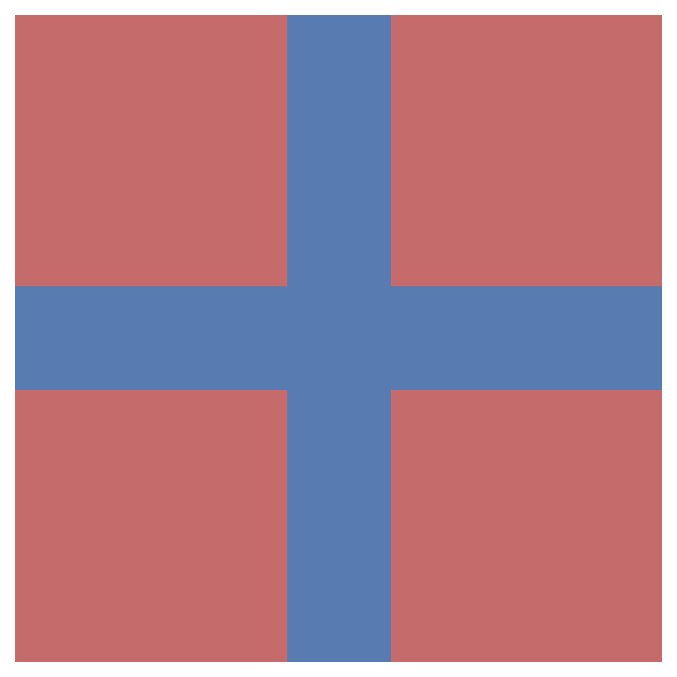}}
  \end{tabular}
  \caption{\textbf{Scaling of Material Expressiveness.} Material distribution matching for 2D torus (top) and cross (bottom). With increasing basis resolution, we find a better match with the target pattern.}
  \label{fig:material_matching}
\end{figure}

\textbf{Material distribution matching.}
To evaluate the material encoding independently of shape variation, we optimize the material score coefficients to reproduce a prescribed spatial pattern on a fixed $50 \times 50$ element grid with $K=2$ materials. Two reference patterns are considered: a 2D torus and a cross. The optimization minimizes a softmax-relaxed label-mismatch loss via CMA-ES at three basis resolutions $N_\Phi \in \{16, 36, 64\}$, corresponding to $4{\times}4$, $6{\times}6$, and $8{\times}8$ RBF grids, see \Cref{fig:material_matching}. At $N_\Phi=16$, the encoder cannot reproduce sharp boundaries. At $N_\Phi=36$, the topology of both patterns is recovered. At $N_\Phi=64$, the match is visually precise. This confirms that finer basis grids translate directly into higher geometric fidelity of the material field.

\setlength{\fboxsep}{0pt}   

\textbf{Intrinsic dimensionality analysis.} 
We uniformly sample $N=2000$ random parameter vectors, decode them into shapes by morphing a reference sphere, and compute pairwise dissimilarities using the symmetric Chamfer distance. Classical multidimensional scaling (MDS) is applied to the resulting distance matrix to obtain a low-dimensional representation~\cite{hout_multidimensional_2013}. Each shape is represented as a point in Euclidean space such that geometric dissimilarities are approximated by Euclidean distances. We define the intrinsic dimensionality estimate $d_{95}$ as the smallest number of MDS components required to explain 95\% of the total variance, providing a measure of the effective dimension of the manifold spanned by the generated shapes.

We compare the proposed basis function encoder against a neural field baseline~\cite{stanley_compositional_2007}. In our comparison, we fix the neural network architecture for each optimization, enabling a controlled evaluation of how representational capacity scales with parameter count. The neural field encoder uses a small multilayer perceptron that maps spatial coordinates to design fields (material scores and displacements), while the network weights and biases serve as the design variables. We test three configurations of increasing capacity --- refining the grid of RBFs and increasing the neural network depth and width --- with approximately matched parameter counts. 

Configurations, parameter counts, and results are reported in \Cref{tab:dimensionality}. For the basis function encoder, $d_{95}$ increases from 70 to 394 as $N_\Psi$ grows. The neural field encoder quickly saturates, with $d_{95}$ increasing marginally from 203 to 238 despite a 20-fold rise in parameters. These findings show that basis functions enable controlled scaling of design complexity, whereas the neural mapping projects onto a high-dimensional manifold largely independent of network size.

\begin{figure}[t]
\centering
\input{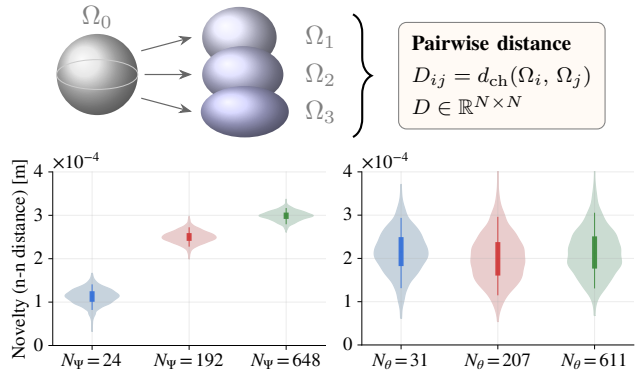}
\caption{\textbf{Basis Functions Expand the Design Space.} Novelty score distribution $\nu_i$ (nearest-neighbor Chamfer distance) for $N{=}2000$ randomly decoded shapes (top) using basis function (left) and neural field encoder (right).}
\label{fig:expressiveness}
\vspace{-1em}
\end{figure}

\textbf{Novelty analysis.}
To complement the dimensionality estimate, we compute a novelty score $\nu_i$ for each decoded shape, defined as its nearest-neighbor Chamfer distance within the sampled set. This measures how dispersed the generated shapes are in the ambient geometry space. As shown in \Cref{fig:expressiveness}, the basis function encoder exhibits steadily increasing novelty with larger $N_\Psi$, whereas the neural field encoder produces nearly identical novelty distributions across architectures. These results reveal two key findings: (i) Increasing novelty with basis refinement indicates a systematically expanding design space versus early saturation in neural fields; and (ii) novelty variance is substantially smaller for the basis function embedding, suggesting more uniform coverage of the design space, while neural fields exhibit more irregular sampling.

\begin{table}[t]
 \centering
 \caption{Intrinsic dimensionality $d_{95}$, the number of principal components capturing 95\% of design variance, for basis function and neural field encoders at three parameter counts.}
  \label{tab:dimensionality}
  \begin{tabular}{llcc}
    \toprule
    Encoder & Config. & Params & $d_{95}$ \\
    \midrule
    \multirow{3}{*}{Basis function} 
      & $2\times2\times2 \times 3$  & 24  & 70  \\
      & $4\times4\times4 \times 3$  & 192 & 244 \\
      & $6\times6\times6 \times 3$  & 648 & 394 \\
      \midrule
    \multirow{3}{*}{Neural field}
      & $3 \to 4 \to 3$              & 31  & 203 \\
      & $3 \to 6 \to 12 \to 6 \to 3$ & 207 & 227 \\
      & $3 \to 8 \to 16 \to 16 \to 8 \to 3$ & 611 & 238 \\
    \bottomrule
  \end{tabular}
\end{table}

\subsection{Co-Optimization of Design and Actuation}
\label{subsec:coopt}

We now evaluate whether jointly optimizing shape, material, and actuation outperforms a sequential strategy in which morphology is fixed before actuation is tuned. We test this on two dynamic tasks, swimming and jumping, under the same total evaluation budget of $10{,}000$ simulator calls. In both tasks, CMA-ES uses $\lambda=50$ over 200 generations. In the \emph{joint} setting, all parameters are optimized together. In the \emph{sequential} setting, shape and material are optimized for 150 generations with fixed actuation, then actuation alone for 50 generations.

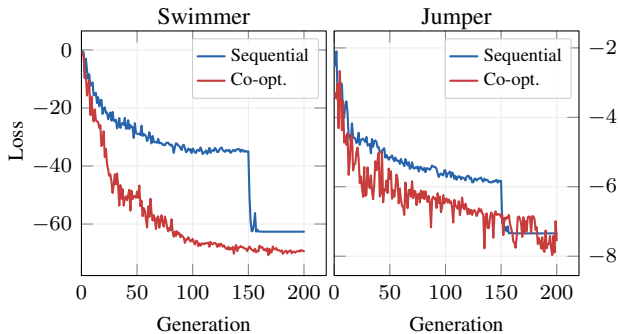
\begin{figure}[!b]
\centering
\vspace{-1em}
\def\datafile{figures/data/opt_loss2.csv}
\def\xcol{generation}
\def\xlabel{Generation}
\def\ylabel{Loss}
\definecolor{plotRed}{RGB}{200, 60, 60}
\definecolor{plotBlue}{RGB}{40, 100, 170}

\def\plotwidth{0.56\columnwidth}

\pgfplotsset{
  compact legend/.style={
    legend style={
      font=\scriptsize,
      at={(0.97,0.97)},
      anchor=north east,
      draw=black!20,
      fill=white,
      fill opacity=0.85,
      text opacity=1,
      rounded corners=0.5pt,
      inner sep=1.5pt,
      row sep=-1pt,
      /tikz/every even column/.append style={column sep=2pt},
    },
    legend image post style={scale=0.6},
    legend cell align=left,
  }
}

\begin{tikzpicture}
  \begin{axis}[
    name=leftplot,
    width=\plotwidth,
    height=\plotwidth,
    title={Swimmer},
    title style={font=\small, yshift=-7pt},
    xlabel={\xlabel},
    ylabel={\ylabel},
    ylabel style={at={(axis description cs:-0.20,0.55)}},
    xmin=0,
    grid=major,
    grid style={black!8, line width=0.2pt},
    tick label style={font=\footnotesize},
    label style={font=\footnotesize},
    line join=round,
    line cap=round,
    compact legend,
  ]
    \addplot[plotBlue, solid, thick] table [col sep=comma, x=\xcol, y=swimmerseq] {\datafile};
    \addlegendentry{Sequential}
    \addplot[plotRed, solid, thick] table [col sep=comma, x=\xcol, y=swimmer] {\datafile};
    \addlegendentry{Co-opt.}
    
  \end{axis}
  \begin{axis}[
    at={(leftplot.east)},
    anchor=west,
    xshift=3pt,
    width=\plotwidth,
    height=\plotwidth,
    title={Jumper},
    title style={font=\small, yshift=-9pt},
    xlabel={\xlabel},
    ylabel={},
    yticklabel pos=right,
    xmin=0,
    grid=major,
    grid style={black!8, line width=0.2pt},
    tick label style={font=\footnotesize},
    label style={font=\footnotesize},
    line join=round,
    line cap=round,
    compact legend,
  ]
    \addplot[plotBlue, solid, thick] table [col sep=comma, x=\xcol, y=jumperseq] {\datafile};
    \addlegendentry{Sequential}
    \addplot[plotRed, solid, thick] table [col sep=comma, x=\xcol, y=jumper] {\datafile};
    \addlegendentry{Co-opt.}
    
  \end{axis}

\end{tikzpicture}%
\caption{\textbf{Optimization Dynamics: Sequential vs Co-Optimization.} Loss over CMA-ES generations for swimmer (left) and jumper (right) under sequential and co-optimization.}
\label{fig:seqcop_lossplots}
\end{figure}

\textbf{Swimming.}
The reference geometry is an ellipsoid with semi-axes $(6,2,2)\,$m, discretized by a tetrahedral mesh. Three materials are used: one passive phase and two longitudinally actuated muscle groups. Muscle assignment is restricted to upper and lower bands to promote bending interactions with the surrounding fluid, which is modeled via a simplified surface-based drag–thrust formulation~\cite{ma_diffaqua_2021}. Actuation consists of squared periodic signals with optimizable frequencies and relative phase. The $(6,2,2)$ RBF grid yields $72$ material, $72$ morphing, and $3$ actuation parameters, totaling $147$ design variables. The objective combines a relative forward displacement reward with penalties for relative lateral drift, rotation, and large muscle fractions:
\begin{equation}
  J_{\text{swim}} = -\alpha_1 \mathcal{L}_{\text{disp}} + \alpha_2 \mathcal{L}_{\text{drift}} + \alpha_3 \mathcal{L}_{\text{rot}} + \alpha_4 \mathcal{L}_{\text{muscle}}.
  \label{eq:swim_loss}
\end{equation}
We use the coefficients $\alpha_1{=}6$, $\alpha_2{=}0.2$, $\alpha_3{=}1$, $\alpha_4{=}1$. The co-optimized swimmer reaches a final loss of $-70.74$ compared to $-62.64$ for the sequential strategy, see \Cref{fig:seqcop_lossplots}. In the sequential run, a visible improvement occurs at generation $150$ when the actuation parameters are freed, but the final performance remains below the joint strategy. The co-optimized swimmer (\Cref{fig:overview}C middle) produces forward propulsion along a stable, nearly straight trajectory. In contrast, the sequential design (\Cref{fig:overview}C bottom) exhibits lateral drift and fails to maintain a consistent swimming direction.

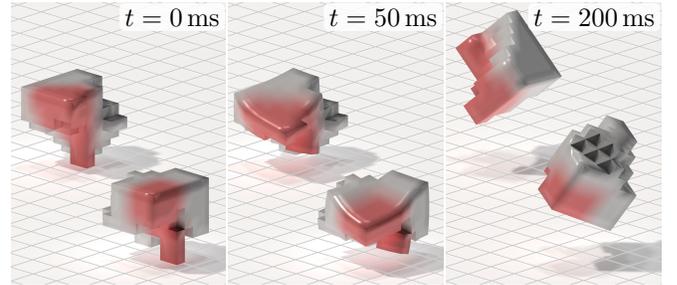
\begin{figure}[t]
 \centering
 \begin{tikzpicture}[baseline=(img.south)]
   \node[inner sep=0, outer sep=0] (img)
     {\includegraphics[width=0.33\columnwidth]{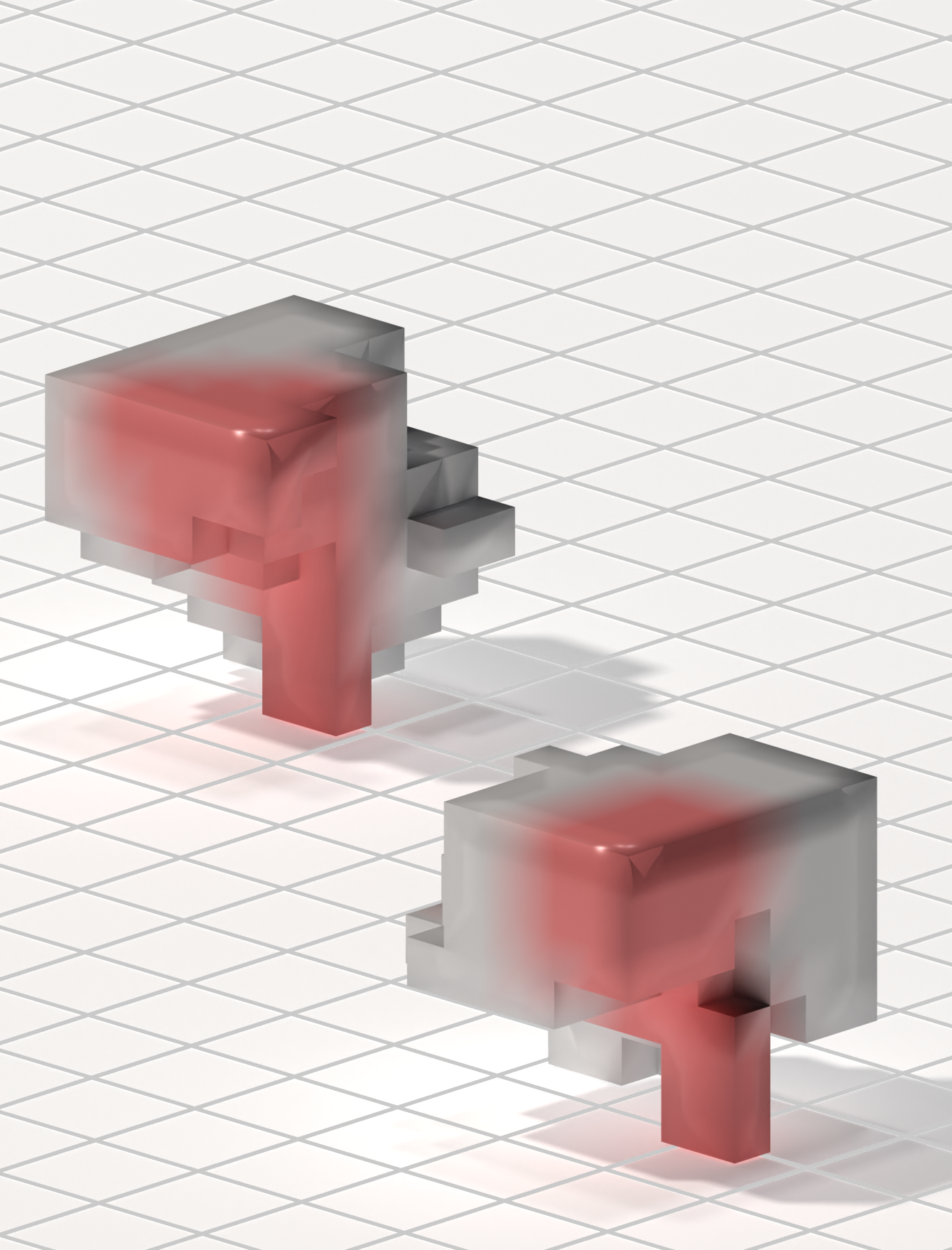}};
   \node[anchor=north east,
          fill=white, fill opacity=0.85, text opacity=1,
          font=\normalsize, rounded corners=2pt,
          inner sep=2pt]
          at (img.north east) {$t=\SI{0}{ms}$};
 \end{tikzpicture}\hfill
 \begin{tikzpicture}[baseline=(img.south)]
   \node[inner sep=0, outer sep=0] (img)
      {\includegraphics[width=0.33\columnwidth]{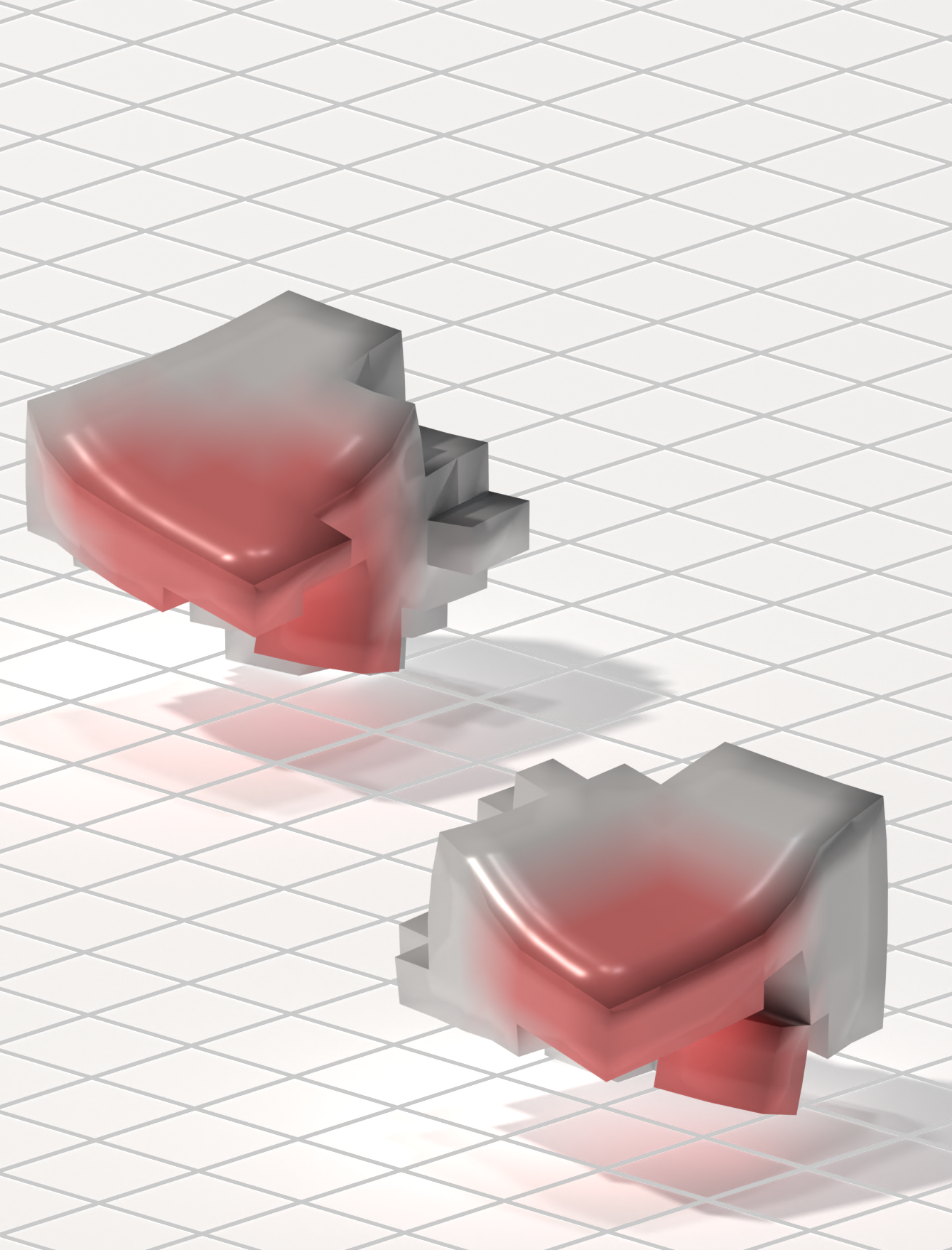}};
   \node[anchor=north east,
          fill=white, fill opacity=0.85, text opacity=1,
          font=\normalsize, rounded corners=2pt,
          inner sep=2pt]
          at (img.north east) {$t=\SI{50}{ms}$};
 \end{tikzpicture}\hfill
 \begin{tikzpicture}[baseline=(img.south)]
   \node[inner sep=0, outer sep=0] (img)
      {\includegraphics[width=0.33\columnwidth]{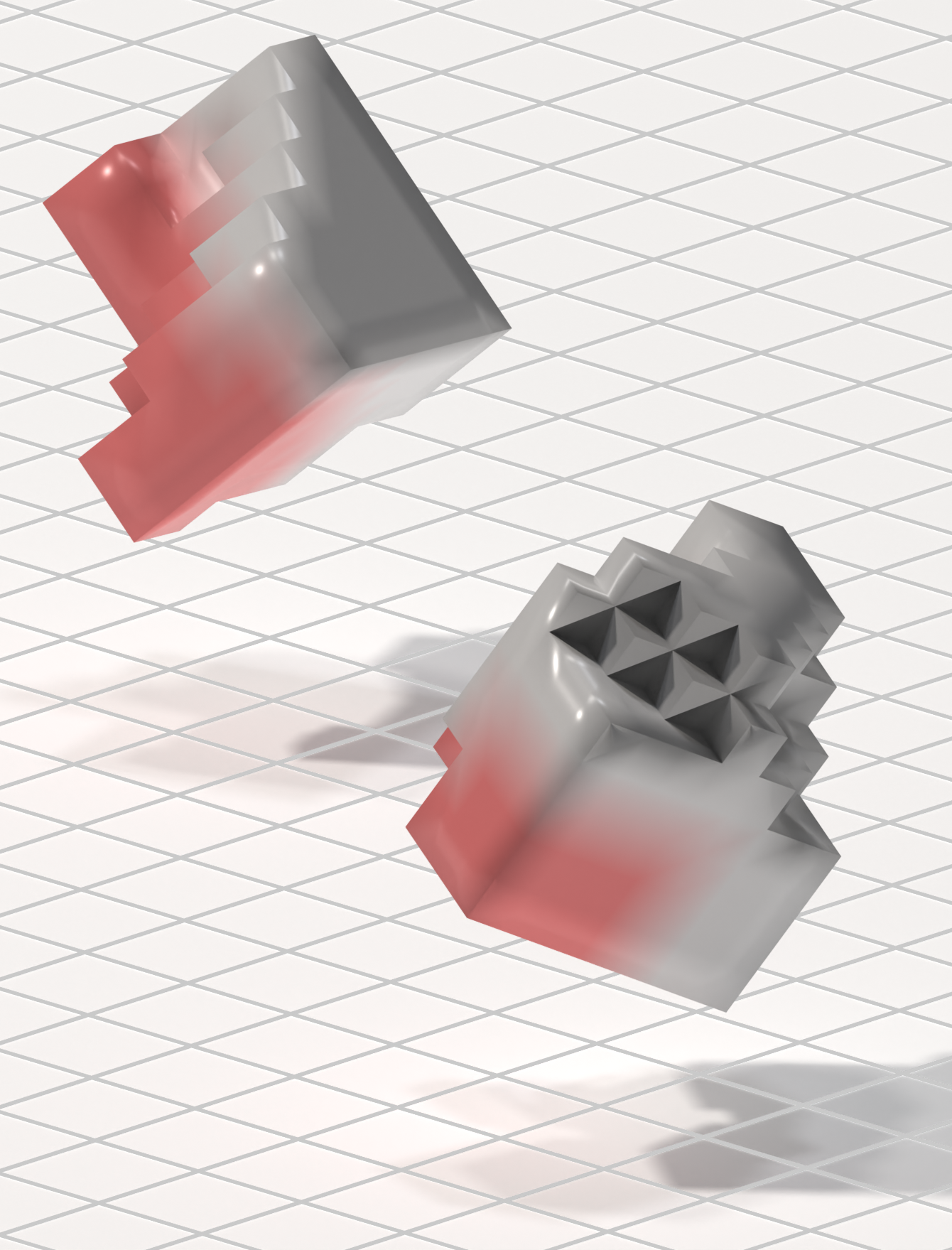}};
   \node[anchor=north east,
          fill=white, fill opacity=0.85, text opacity=1,
          font=\normalsize, rounded corners=2pt,
          inner sep=2pt]
          at (img.north east) {$t=\SI{200}{ms}$};
 \end{tikzpicture}
 \caption{\textbf{Co-Optimization Yields Superior Jump Dynamics.} Snapshots at rest, ground-contact compression, and mid-flight for co-optimized (front) and sequentially optimized (back) jumper designs.}
 \label{fig:three_jumpers}
 \vspace{-1em}
 \end{figure}

\textbf{Jumping.}
The second task considers vertical jumping with rotation under gravity and ground contact. The reference geometry is a cube of side $\SI{0.1}{m}$, discretized into a $7 \times 7 \times 7$ hexahedral grid ($N_e=343$). Two materials are used: a passive elastic phase and a vertically actuated muscle phase driven by a Gaussian pulse with optimizable peak time, amplitude, and width ($N_a=3$). Shape variation occurs solely through topology changes via element removal, without morphing. The design comprises $54$ material and $3$ actuation parameters ($N_c = 57$). The objective rewards both relative jump height and accumulated rotation, and penalizes large muscle fractions:
\begin{equation}
  J_{\text{jump}} = -\beta_1 \mathcal{L}_{\text{jump}} - \beta_2 \mathcal{L}_{\text{rot}} + \beta_3 \mathcal{L}_{\text{muscle}}.
  \label{eq:jump_loss}
\end{equation}
The coefficients are $\beta_1{=}12$, $\beta_2{=}1$, $\beta_3{=}1$. The co-optimized jumper reaches a final loss of $-7.97$ versus $-7.34$ for the sequential strategy, again showing a late improvement once actuation is released (see \Cref{fig:seqcop_lossplots}). \Cref{fig:three_jumpers} shows trajectory snapshots for both designs at three key time instants: initial configuration, ground-contact compression, and mid height. Both designs converge to asymmetric geometries with non-trivial element removal patterns, while the co-optimized solution achieves even faster rotation.

\subsection{Comparison to Baseline Encodings}
We compare the proposed basis function encoding against two alternative design representations: the neural field encoding introduced in the dimensionality analysis on the swimming task and a per-voxel encoding on the jumping task. In both cases, all optimization settings are identical to the sequential vs. co-optimization experiments.

\textbf{Neural field encoding for swimming.}
The neural baseline uses two fully connected networks for material ($3\to8\to3$, $59$ parameters) and morphing ($3\to12\to3$, 87 parameters). With $3$ actuation parameters, this gives $N_c^{\text{neural}}=149$, comparable to $N_c=147$ for the basis function model. Under the original objective from the swimmer co-optimization experiment, the neural field encoder consistently converged to a stationary solution with no muscle elements. To prevent this, the forward displacement reward weight was doubled for the neural baseline. For a fair comparison, the basis function encoder was also re-evaluated under the modified objective. Despite the neural field being differentiable in principle, both encoders are optimized with CMA-ES to isolate representation effects from optimizer choice. 

\begin{figure}[t]
\centering
\input{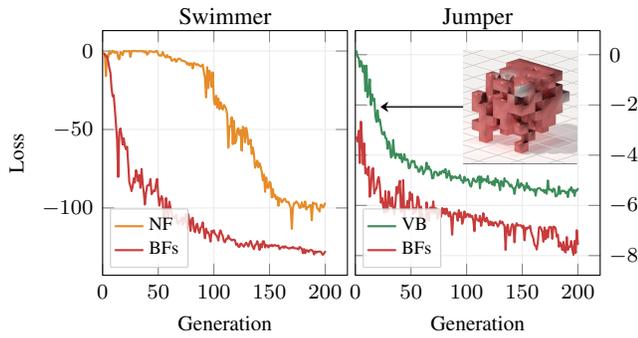}
\caption{\textbf{Basis Functions Outperform Alternative Encodings.} Task loss for the swimmer (left) using neural field vs.\ basis function encoding, and the jumper (right) using voxel-based vs.\ basis function encoding.}
\label{fig:baselines_lossplots}
\vspace{-1em}
\end{figure}

\Cref{fig:baselines_lossplots} shows that the neural field encoder makes no progress during roughly the first half of optimization. In contrast, the basis function encoder improves strongly already within the first few generations. Since the basis function design and trajectory under the modified loss are visually similar to the original (\Cref{fig:overview}), they are not shown separately. The neural field design achieves net forward displacement but exhibits visible lateral drift. The key take-away, next to performance, is the clear difference in the loss landscape: the linear, spatially local basis parameterization yields a more navigable objective than the globally coupled neural encoding, where each weight influences the entire design through nonlinear compositions.

\textbf{Per-voxel encoding for jumping.}
The per-voxel baseline assigns one material variable per element ($N_e=343$) plus 3 actuation parameters, yielding $N_c^{\text{voxel}}=346$, roughly $6\times$ larger than the basis function configuration ($N_c=57$). The per-voxel encoding does not support shape morphing, matching the basis function setup for this task. All other optimization settings are identical. Both encodings find jumping solutions, but the basis function encoder reaches a better final loss despite using $6\times$ fewer parameters. \Cref{fig:baselines_lossplots} shows the loss plots and the optimized per-voxel design. The per-voxel encoder does not benefit from its additional parameter count: the element-wise parameterization yields a much larger, weakly regularized search space that is harder to explore under a fixed evaluation budget. The basis function jumper displays coherent, compact muscle regions, whereas the per-voxel design yields a fragmented distribution with scattered muscles and a higher muscle fraction. While functional, such a design would be difficult to realize in practice.

\section{Conclusion}

This work introduced a unified, low-dimensional design embedding for soft robot optimization that jointly parameterizes shape, multi-material distribution, and actuation within a single structured parameter space. We find that our embedding provides predictable and tunable control over design space complexity by showing that the intrinsic dimensionality of the induced design manifold scales systematically with the number of basis functions. In contrast, the neural field encoder shows little change in intrinsic dimensionality as parameters increase. We further show that joint co-optimization of morphology and actuation consistently outperforms sequential strategies under identical evaluation budgets. Lastly, the basis function representation yields a more navigable objective landscape than neural fields while enforcing spatial coherence and lower dimensionality than per-voxel parameterizations. Overall, structuring the design space --- rather than modifying the simulator or relying on differentiability --- enables effective black-box co-design.

Several limitations should be noted. Multi-objective losses require manual weighting, actuation is open-loop, and fabrication constraints are not explicitly enforced but handled implicitly by the simulator. Future work will focus on extending the framework beyond open-loop black-box optimization. Integrating reinforcement learning or policy-gradient methods would enable state-dependent control and the co-design of morphology with feedback policies. Further, adaptive basis function placement and refinement strategies could be investigated to overcome the limitations of the currently fixed uniform grid. The differentiable decoding map could be paired with a differentiable simulator or learned surrogate for gradient-based optimization. Adding manufacturability objectives and physics-aware regularization would promote sim-to-real transfer. Experimental fabrication and validation remain key steps toward translating computational co-design into physical soft robotic systems.

\printbibliography
\end{document}